%% file: main_ieee.tex
\documentclass[journal]{IEEEtran}

\usepackage{balance} 
\usepackage{graphicx}
\usepackage{subfigure}
\usepackage{float}
\usepackage{amsfonts}
\usepackage{enumerate}
\usepackage{algorithm}
\usepackage{algpseudocode}
\usepackage{amsmath}
\usepackage{enumitem}
\usepackage{caption}
\usepackage{hyperref}
\usepackage[marginal]{footmisc}

\begin{document}

\title{DeepFreight: Integrating Deep Reinforcement Learning and Mixed Integer Programming for Multi-transfer Truck Freight Delivery}
\author{Jiayu Chen, Abhishek K. Umrawal, Tian Lan, and Vaneet Aggarwal \thanks{J. Chen, A. U. Umrawal, and V. Aggarwal are with Purdue University, West Lafayette IN 47907, USA, email: \{chen3686,aumrawal,vaneet\}@purdue.edu. T. Lan is with the George Washington University, Washinton DC 20052, USA, email:tlan@gwu.edu.  

This work was presented in part at ICAPS 2021. 
}}
\maketitle
\input{abstract}

\input{intro}

\input{model}

\input{algo}

\input{milp}

\input{evaluations}

\input{conclusion}

\bibliographystyle{IEEEtran} 
\bibliography{ref}

\end{document}

%% file: abstract.tex
\begin{abstract}
With the freight delivery demands and shipping costs increasing rapidly, intelligent control of fleets to enable efficient and cost-conscious solutions becomes an important problem. In this paper, we propose DeepFreight, a model-free deep-reinforcement-learning-based algorithm for multi-transfer freight delivery, which includes two closely-collaborative components: truck-dispatch and package-matching. Specifically, a deep multi-agent reinforcement learning framework called QMIX is leveraged to learn a dispatch policy, with which we can obtain the multi-step joint vehicle dispatch decisions for the fleet with respect to the delivery requests. Then an efficient multi-transfer matching algorithm is executed to assign the delivery requests to the trucks. Also, DeepFreight is integrated with a Mixed-Integer Linear Programming optimizer for further optimization. The evaluation results show that the proposed system is highly scalable and ensures a 100\% delivery success while maintaining low delivery-time and fuel consumption. The codes are available at \href{https://github.com/LucasCJYSDL/DeepFreight}{https://github.com/LucasCJYSDL/DeepFreight}.

\end{abstract}


%% file: intro.tex
\section{Introduction}
According to American Trucking Associations\footnote{\href{https://www.trucking.org/economics-and-industry-data}{https://www.trucking.org/economics-and-industry-data}}, the U.S. trucking industry shipped 11.84 billion tons of goods in 2019 and had a market of 791.7 billion dollars. With shipping costs rising and freight volumes outpacing the supply of available trucks, companies are thinking of radical new initiatives to get their products into customers' hands more efficiently. On the other hand, transportation is currently the largest source (29\%) of greenhouse gas emissions in the U.S., where the passenger cars and trucks account for more than 80\% of this section\footnote{\href{https://www.climatecentral.org/gallery/graphics/emissions-sources-2020}{https://www.climatecentral.org/gallery/graphics/emissions-sources-2020}}. Thus, innovations to enable intelligent and efficient freight transportation are of central importance for the sake of better utilization of the trucks and less fuel consumption.

We propose DeepFreight, a model-free learning framework for the freight delivery problem. 
The proposed algorithm decomposes the problem into two closely-collaborative components: truck-dispatch and package-matching. In particular, the dispatch policy aims to find the dispatch decisions for the fleet according to the delivery requests. It leverages a multi-agent reinforcement learning framework, QMIX \cite{rashid2018qmix}, which can train decentralised policies in a centralised end-to-end fashion. Then an efficient matching algorithm is developed to match the delivery requests with the dispatch decisions, which is a greedy approach that makes use of multiple transfers and aims at minimizing the total time use of the fleet. Further, DeepFreight is integrated with a Mixed-Integer Linear Programming (MILP) optimizer for more efficient and reliable dispatch and assignment decisions.

To the best of our knowledge, this is the first work that takes a machine learning aided approach for freight scheduling with multiple transfers. The key contributions of the paper can be summarized as follows:


\begin{enumerate}[leftmargin=*]
    \item We propose DeepFreight, a learning-based algorithm for multi-transfer freight delivery, which contains truck-dispatch and package-matching. The dispatch policy is adopted to determine the routes of the trucks, and then matching policy is executed to assign the packages to the trucks efficiently. 
    \item We propose to train the dispatch policy through a centralized training with decentralized execution algorithm, QMIX \cite{rashid2018qmix}, which considers the cooperation among the trucks when training and makes the multi-agent dispatch  scalable for execution.
    \item We propose an efficient rule-based matching policy which allows multi-transfer to minimize the usage time of the fleet.
    \item We formulate a MILP model for this freight delivery problem and integrate it with DeepFreight to achieve better reliability and efficiency of the overall system.

\end{enumerate}

The rest of this paper is organized as follows. Section \ref{sec:related} introduces some related works and highlights the innovation of this paper. Section \ref{sec:system} presents an example for the multi-transfer freight ride-sharing system and lists the model parameters and optimization objectives of the system for later use. Section \ref{sec:algo} explains the proposed approach: DeepFreight in details, including its overall framework and main algorithm modules. Section \ref{sec:milp} introduces a hybrid approach that harnesses DeepFreight and MILP: DeepFreight+MILP, including the formulation of MILP and the workflow of this hybrid approach. Section \ref{sec:eval} talks about evaluation and results, including the simulation setup, implementation details of DeepFreight and comparisons among the algorithms: DeepFreight, DeepFreight without Multi-transfer, MILP and DeepFreight+MILP. Section \ref{sec:conc} concludes the paper with a discussion of future work.

\clearpage

\section{Related Work}\label{sec:related}

{\bf  Truck Dispatch Problem.} The freight delivery problem is a variant of `Vehicle Routing Problem (VRP)', which was first introduced in \cite{dantzig1959truck} as the `Truck Dispatch Problem' and then generalized in \cite{clarke1964scheduling} to a linear optimization problem: how to use a fleet of trucks with capacity constraints to serve a set of customers that are geographically dispersed around the central depot. Over the decades, the classic VRP has been extended to a lot of variants by introducing additional real-life characteristics \cite{braekers2016vehicle}. For example, VRP with Pickup and Delivery \cite{parragh2008survey}, that is, goods are required to be picked up and dropped off at certain locations and the pick-up and drop-off must be done with the same vehicle; VPR with Time Windows \cite{dumas1991pickup}, that is, the pickups or deliveries for a customer must occur in a certain time interval; Multi-Depot VRP \cite{montoya2015literature}, which assumes that there are multiple depots spreading among the customers for scheduling. The extended VRP is NP-hard, and many heuristics and meta-heuristics algorithms have been adopted as the solution. Among them, the meta-heuristics algorithms are used the most \cite{braekers2016vehicle}, such as simulated annealing \cite{wang2015parallel}, genetic algorithm \cite{tasan2012genetic}, particle swarm optimization \cite{goksal2013hybrid}, and local search algorithm \cite{avci2015adaptive}. 

We note that these works don't consider that goods can be dropped off in the middle and carried further by another truck (multi-transfer of goods), which has the potential for better utilization of the fleet and is considered in our work. Further, the approaches they adopt are model-based, which need execution every time new observation occurs. Also, the execution time would increase with the scale of the problem, so they usually have high time complexity and poor scalability.

{\bf  Multi-hop Ride-Sharing.} As a related area, researches about the ride-sharing system are introduced in this part. To handle the dynamic requests in this scenario, reinforcement learning aided approaches for efficient vehicle-dispatch and passenger-matching have been proposed in \cite{oda2018movi,al2019deeppool,de2020efficient}. However, these approaches don't consider passengers going over multiple hops (transfers). As shown in \cite{teubner2015economics}, the multi-transfer has the potential to greatly increase the ride availability and reduce emissions of the ride-sharing system by making better use of the vehicle capacities. Given this, a reinforcement-learning based approach for the multi-hop ride-sharing problem has been recently studied in \cite{singh2019distributed}. 

However, we note that there are some key differences between the ride-sharing system and freight-scheduling system. For ride-sharing, the driving distances are smaller and it's more concerned with the real-time scalable solutions for the highly dynamic requests. While for freight-scheduling, the number of the trucks needed is much fewer due to their huge capacity and the delivery demands can usually be acquired in advance (e.g. one day ahead). In this case, the coordination among trucks is possible and necessary for making better scheduling decisions and we try to take it into consideration with our freight delivery system through a multi-agent reinforcement learning setting.

{\bf Multi-Agent Reinforcement Learning.} Multi-agent reinforcement learning (MARL) methods hold great potential to solve a variety of real-world problems, and centralized training with decentralized execution (CTDE) is an important paradigm of it. Centralized training exploits the interaction among the agents, while decentralized execution ensures the system's scalability. Specific to the multi-agent cooperative setting (like ours), there are many related works: VDN \cite{sunehag2018value}, QMIX \cite{rashid2018qmix}, QTRAN \cite{son2019qtran} and MAVEN \cite{mahajan2019maven}. However, their performance varies greatly in different benchmarks and none of them has the absolute advantage. Considering there have been more applications of QMIX in complicated scenarios and its robust performance \cite{zhang2020dynamic,iqbal2020ai}, we choose QMIX to train the dispatch policy in our algorithm framework.

%% file: model.tex
\section{System Model}\label{sec:system}
\begin{figure}[t]
    \centering
    \includegraphics[width=3.3in,height=1.5in]{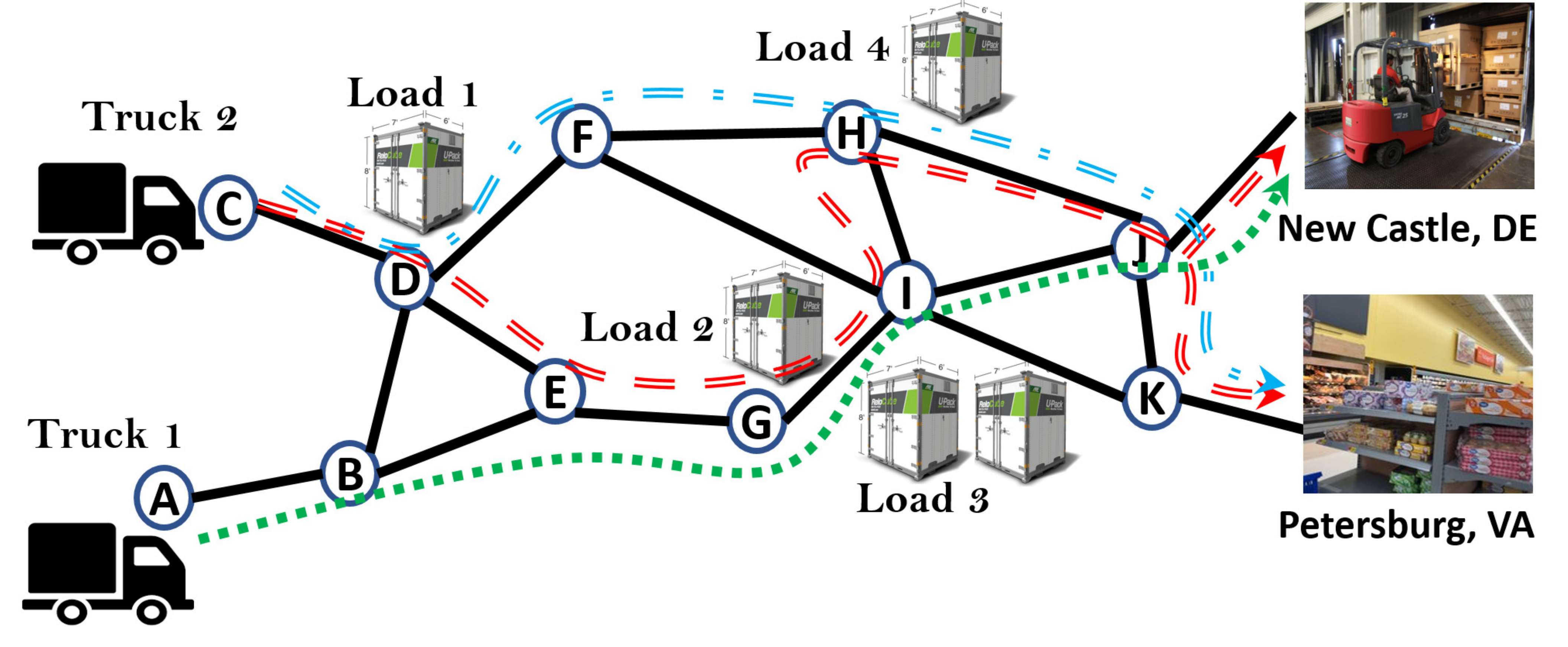}
    \caption{An example for multi-transfer freight delivery}
    \label{1}
\end{figure}

In this section, we will describe the proposed system model for multi-transfer freight delivery. As an example, consider the scenario shown in Figure \ref{1}, where Loads 1 and 4 need to be shipped to a distribution center in Petersburg, VA, while Loads 2 and 3 to the distribution center in New Castle, DE. There are two trucks located at zone A and zone C, with capacity $R_{1} = 4$ and $R_{2} = 5$, respectively. When freight ride-sharing is enabled, two different possibilities are shown in the figure to serve the requests: 1) serving all the requests using only the truck 2 at zone C (i.e., the dashed red line in the figure), 2) serving Loads 1 and 4 using truck 2 (i.e., the dashed-dotted blue line) and Loads 2 and 3 using truck 1 (i.e., the dotted green line), offering different tradeoffs between delivery time and fuel cost. Furthermore, if multiple transfers are allowed, we may also have the two trucks pick up their loads first and then meet at zone H to fold the loads into a single truck before the final delivery, which provides additional elasticity for the truck and freight scheduling problem. Through adoption of freight ride-sharing and multi-transfer, fewer truck miles may be required, which will lead to reduced shipping costs and less fuel consumption.

\subsection{Model Parameters}
In this section, key parameters of the system model are introduced. First, there are $N_s$ distribution centers in the freight delivery system, which can be used as the sources and destinations of packages. Based on them, a delivery request list, denoted by $R_{1:N_r}$ ($N_r$: the number of requests), is collected at the beginning of each day. The $i$-th request $r_i$ can be represented as ($source_i$, $destination_i$, $size_i$), which indicates the source, destination, and size of the $i$-th package.

Second, our target is to complete these requests using a fleet of trucks within a time limit. The number of the trucks for dispatch is $N_{t}$ and the time limit is denoted as $T_{max}$ which can be one or two days. To fulfill this target, we need to schedule the itinerary of the fleet and the assignment of the packages to the trucks. 

Further, in our system, the itinerary of a truck is not decided as a whole and instead consists of a sequence of dispatch decisions, each of which shows the truck's next stop (distribution center) to visit. Also, the assignment of the packages allows multiple transfers, which means that a package can be assigned to more than one truck along the route from its origin to its destination. Note that the total driving time of a truck can't exceed the time limit $T_{max}$ and the volume of the packages it takes at the same time should be within its maximum capacity $C_{max}$.

\subsection{Problem Objectives}\label{subsec:obj}
The key objectives of the freight delivery system are: 
\begin{enumerate}
    \item Maximizing the number of served requests $N_{rs}$ within the time limit $T_{max}$;
    \item Minimizing the total fuel consumption of the fleet $F_{total}$ during this process.
\end{enumerate}
The second component $F_{total}$ is viewed as the function of the total driving time (denoted as $T_{drive}$) and can be calculated with Equation (1), where $fuel\_factor$ is a constant, representing the fuel consumption per unit time, while $eta_{j}^{k}$ is the estimated time of arrival (ETA) for the $k$-th dispatch decision of truck $j$ that can be obtained through Google Map API. Within an episode (one or two days), every truck will receive a dispatch decision at each epoch, which denotes the next stop to visit. There are at most $N_{e}$ epochs within an episode.
\begin{equation}
    F_{total} =fuel\_factor*T_{drive}
\end{equation}
\begin{equation}
    T_{drive} =\sum_{k=1}^{N_e}\sum_{j=1}^{N_t}eta_{j}^{k}
\end{equation}

%% file: algo.tex
\begin{figure}[t]
	\centering
	\includegraphics[width=3.5in,height=2.2in]{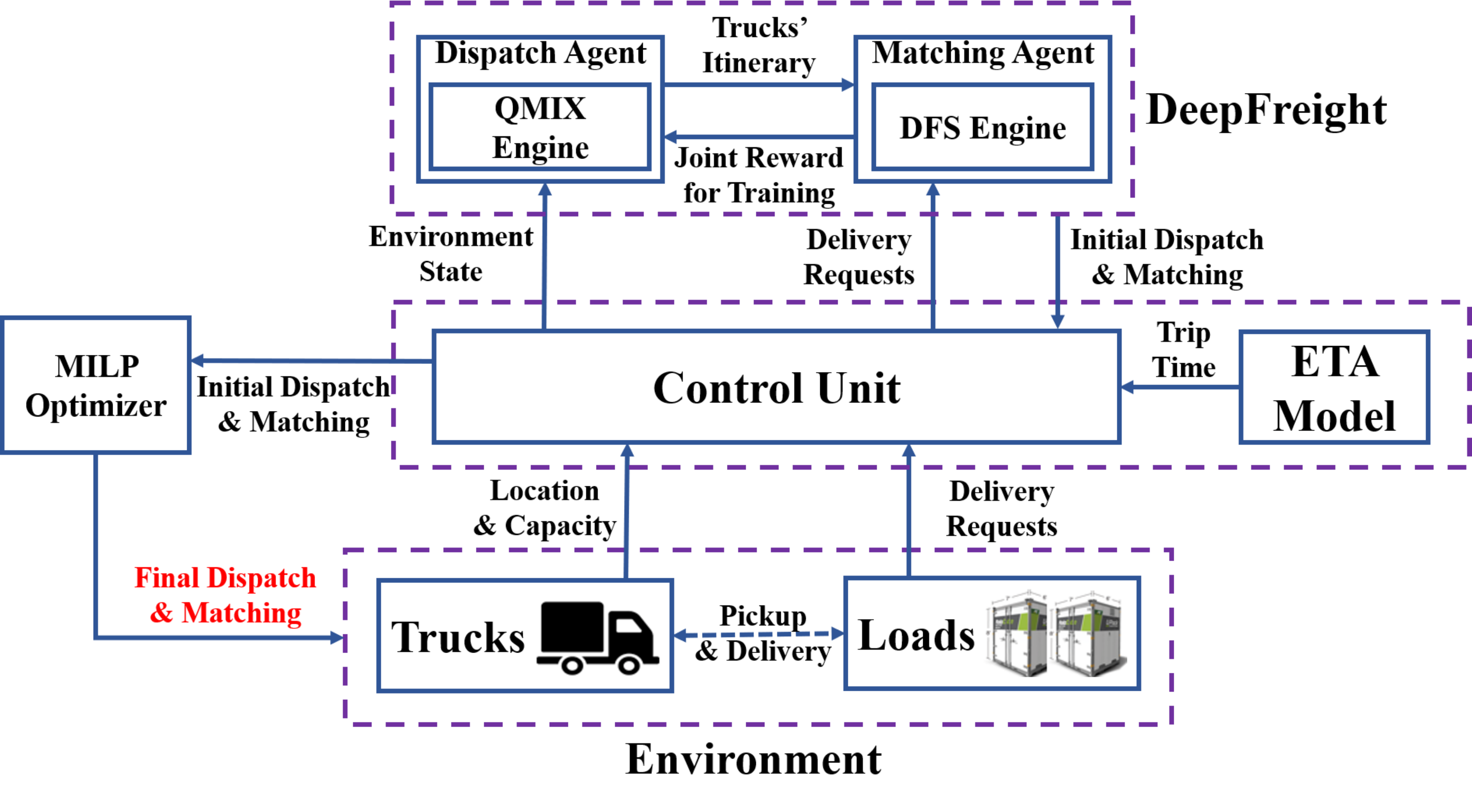}
	\caption{System Structure. DeepFreight (introduced in Section \ref{sec:algo}) decomposes the freight delivery problem into two components: truck-dispatch and package-matching. The dispatch policy determines the itinerary for each truck, which is trained through QMIX. The matching policy is then executed to assign the requests to the trucks based on the route of each truck through DFS. Further, a MILP solver is adopted to optimize the initial dispatch \& matching decisions and feedback the final decisions to the environment (introduced in Section \ref{sec:milp}).}
	\label{0}
\end{figure}

\section{Proposed Approach: DeepFreight}\label{sec:algo}

In this section, we introduce our proposed approach: DeepFreight Algorithm. First, we describe its overall framework in Section \ref{subsec:design}, including the main algorithm modules and connections between them. Then, in Section \ref{subsec:pomdp} -\ \ref{subsec:match}, we talk about these modules further, including the detailed procedure and the intuition to propose them.

\subsection{Overall Framework of DeepFreight}\label{subsec:design}
\textbf{Algorithm 1} shows the overall framework of DeepFreight. As mentioned, the proposed approach should give out the itinerary of the fleet and the assignment of the packages to the trucks. In this case, as shown in Figure \ref{0}, we split the whole system into two parts: the dispatch policy and matching policy. The dispatch policy determines a sequence of dispatch decisions for each truck in the fleet according to the unfinished delivery requests, based on which we can get the route network of the fleet. The matching policy is then executed to assign the requests to the trucks based on the route of each truck. We note that the matching policy is executed based on the dispatch results, while then the matching results provide reward feedback for the training of the dispatch policy. The dispatch policy is trained in a multi-agent reinforcement learning setting called QMIX, which includes two key components: the agent network $Q_{single}$ and mixing network $Q_{mix}$ (introduced in Section \ref{subsec:qmix}).

There are some key points to note about \textbf{Algorithm 1}. First, Line 3-6 is the initialization process of the simulator. A new request list will be generated at the beginning of each episode, however, the truck list will not be generated again until the next operation cycle starts. Within an operation cycle (including several episodes in a row), the trucks' locations at the beginning of a new episode are the same as those at the end of the previous episode. This connection in locations should be taken into consideration when making dispatch decisions. Also, we note that the generation of the request list and truck list is model-based, which is introduced in Section \ref{subsec:setup}.

Second, Line 7-17 shows the sampling process of the agents which is used to collect the experiences for training. In order to prune the joint search space of the agents and further improve the learning efficiency, which is essential for solving this complex problem, we add some constraints to the truck's routes: 1) It's not allowed for the truck to pass through the same stop twice, except for returning to its origin (no sub-tours); 2) No dispatch decisions are allowed any more after the truck returns to its origin or its cumulative time exceeds the time limit.  These constraints are realized through action mask $m_{1:N_{t}}$: once a stop has been passed through by truck $i$, the corresponding term in $m_{i}$ will be set as 1, representing the corresponding dispatch decision (action) is not valid any more.

Third, not all the dispatch decisions can be matched with delivery requests. For a truck, there may be some idle route segments with no package assignment and these idle dispatch decisions (route segments) can be eliminated on the condition that the truck's route continuity is not affected (show in Line 18). Through this, total fuel consumption is further reduced and better scheduling of the fleet is realized.

\begin{algorithm}[t]
	\caption{DeepFreight Algorithm}
        {
			\begin{algorithmic}[1]
				\State \textbf{Given}: episode time limit $T_{max}$, operation cycle $\Delta$, agent network $Q_{single}$, mixing network $Q_{mix}$
				\For {episode $epi=1$ to $N_{epi}$}
				\State Generate $N_{r}$ delivery requests randomly
				\If{$epi$ \% $\Delta$ == 1}
				\State Generate $N_{t}$ trucks randomly
				\EndIf 
				\For {epoch $k=1$ to $N_{e}$}
				\State Obtain the current environment state $s^{k}$, obser-
				\Statex $\qquad \quad$vation list $z_{1:N_{t}}^{k}$, and action mask list $m_{1:N_{t}}^{k}$
				\State Get the q-value list $Q_{1:N_{t}}^{k}$ from $Q_{single}$
				\State Choose the available action $u_{1:N_{t}}^{k}$ for each truck
				\Statex $\qquad \quad$based on the corresponding $Q_{1:N_{t}}^{k}$ and $m_{1:N_{t}}^{k}$, 
				\Statex $\qquad \quad$using Boltzmann distribution policy \cite{cesa2017boltzmann}
				\State Match the delivery requests with the dispatch 
				\Statex $\qquad \quad$decisions using \textbf{Algorithm 2} 
				\State Calculate joint reward $r^{k}$ based on Equation (3)
				\State Update the cumulative time for each truck $t_{1:N_{t}}^{k}$
				\If{$min(t_{1:N_{t}}^{k})\geq T_{max}$ \textbf{or} no requests left}
				\State \textbf{break}
				\EndIf
				\EndFor
				\State Eliminate the useless dispatch decisions and calcu-
				\Statex $\quad \ $ late the unfinished package number and average 
				\Statex $\quad \ $ driving time of the fleet for evaluation
				\State Update replay buffer $B$ with experiences collected 
				\Statex $\quad \ $ above, including ($s^{1:N_{e}}$, $z_{1:N_{t}}^{1:N_{e}}$, $u_{1:N_{t}}^{1:N_{e}}$, $r^{1:N_{e}}$, $m_{1:N_{t}}^{1:N_{e}}$)
				\For{iteration $iter=1$ to $N_{iter}$}
				\State Sample a random batch of experiences from $B$ 
				\State Update $Q_{single}$ and $Q_{mix}$ by minimizing loss
				\Statex $\qquad \quad$function defined as Equation (5)
				\EndFor
				\EndFor
	\end{algorithmic}}
\end{algorithm}

\begin{figure*}[!t]
	\centering
	\includegraphics[width=6.6in, height=1.4in]{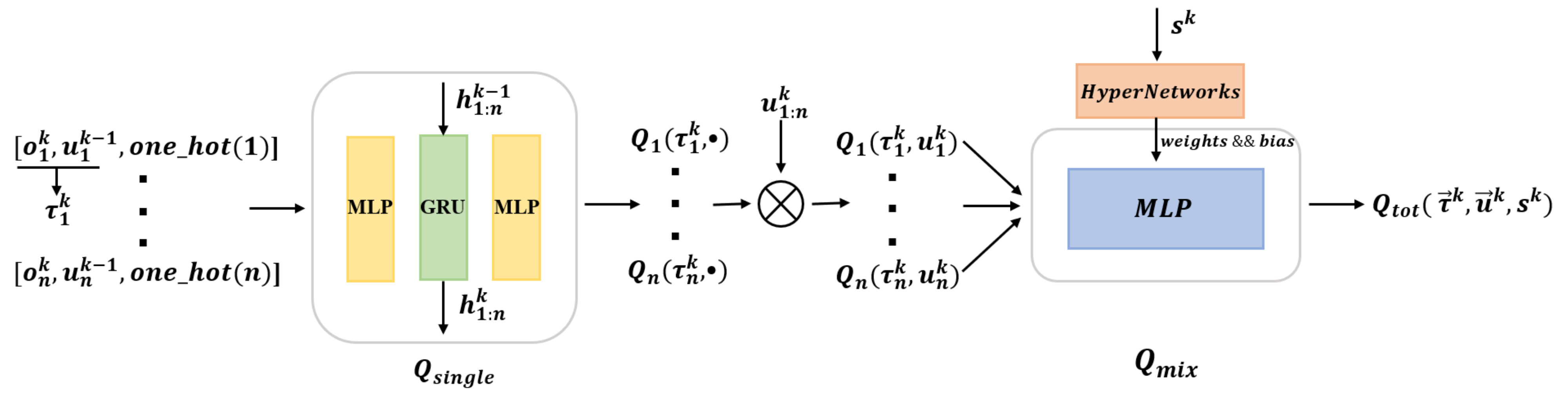}
	\caption{Workflow of QMIX Algorithm}
	\label{2}
\end{figure*}

Finally, some training tricks about deep Q-Learning are adopted. For example, Boltzmann distribution \cite{cesa2017boltzmann} is used to improve the exploration of the environment and the experience replay buffer is used to improve the efficiency of sample use and the stability of training. 

\subsection{Modeling Dispatch problem with Decentralized Partial Observable Markov Decision Process (Dec-POMDP)}\label{subsec:pomdp}
As a cooperative multi-agent task, the dispatch problem is modeled with Dec-POMDP, which can be described as a tuple $(S, U, P, r, Z, O, n, \gamma)$. The general definition of the elements in the tuple are as follows:
\begin{itemize}
	\item $n$ : the number of agents used in the task;
	\item $S$ : the environment state space;
	\item $U$ : the individual action space;
	\item $\mathbf{U} \equiv U^n$: the joint action space for the n agents;
	\item $P(s{'}|s, \mathbf{u})$ : $S \times \mathbf{U} \times S \to [0, 1]$, the state transition function;
	\item $r(s, \mathbf{u})$ : $S \times \mathbf{U} \to \mathbb{R}$, the reward function based on and shared by all the agents;
	\item $Z$ : the individual observation space;
	\item $O(s, a)$ : $S \times A \to Z$, the observation function that each agent in the list $A=\{1,2,\cdot\cdot\cdot,n\}$ draws individual observation $z\in Z$ according to;
	\item $\gamma \in [0, 1)$ : the discount factor. 
\end{itemize}

Specifically, the environment state space $S$ for this package delivery dispatch problem includes four terms: 
\begin{enumerate}
    \item current epoch number $k$;
    \item distribution of the trucks at epoch $k$: the number of the trucks at each stop;
    \item available capacity of each truck;
    \item unfinished delivery requests at epoch $k$, which are initialized at the beginning of each episode and updated with the epochs.
\end{enumerate}

As for the individual observation space $Z$, it is for a single truck, so it replaces the item 2) 3) with the truck's current stop and available capacity respectively and keeps item 1) 4). 

Moreover, the individual action space $U$ includes:
\begin{enumerate}
    \item $next\_stop \in \{1, 2, \cdot\cdot\cdot, N_s\}$ : the index of the next stop;
    \item invalid dispatch: there are some constraints for the dispatch decisions to be valid (e.g. within the time limit).
\end{enumerate}
In this case, the dimension of the individual action space $U$ is $N_s+1$.

The joint reward function $r$ should reflect the objectives mentioned in Section \ref{subsec:obj}, so the reward for the epoch $k$ is defined as Equation (3), where $\beta_{1:2} > 0$.
\begin{equation}
r_{k}=\beta_{1}N_{rs}^{k} - \beta_{2}F_{total}^{k}
\end{equation}
Obviously, it has a positive correlation with the package number served at epoch $k$ and a negative correlation with the fuel consumption during this process.

Based on these elements, the dispatch policy  for agent $a$ can be defined as $\pi^{a}(u^{a}|z^{a}) : Z \times U \to [0, 1]$. The execution of the policy is decentralized, since it conditions only on the observation of the single agent. However, the training process is centralized through maximizing the joint Q function for the joint dispatch policy $\pi$, which is defined as Equation (4).
\begin{equation}
Q^{\pi}(s_{t}, \mathbf{u}_{t}) =\mathbb{E}[\sum_{i=t}^{T}\gamma^{i-t}r_{i}]=\mathbb{E}[\sum_{i=t}^{T}\gamma^{i-t}r(s_{i}, \mathbf{u}_{i})]
\end{equation}

\subsection{QMIX for Training the Dispatch Policy}\label{subsec:qmix}

QMIX \cite{rashid2018qmix} is a value-based reinforcement learning algorithm that can train decentralized policies in a centralized end-to-end fashion. It is suitable to solve this multi-agent freight delivery problem because: 1) the centralized training process takes the coordination among trucks into consideration; 2) the decentralized execution process alleviates the complexity to determine the joint dispatch decisions and ensures the scalability of the dispatch policy. The workflow of QMIX is shown as Figure \ref{2}, where two key components: $Q_{single}$ and $Q_{mix}$, need special attention.

The agent network $Q_{single}$ outputs the Q-value list for each truck, which can be used for decentralized execution, or as a part of the centralized training process. The inputs of $Q_{single}$ include: 1) the agent (truck) id $a$, which is in the form of a one-hot vector; 2) the individual observation for the current epoch $o_{a}^{k}$; 3) the dispatch decision (action) for the previous epoch $u_{a}^{k-1}$. Input 1) ensures that the dispatch policy is unique for each truck, which means even if different trucks have the same observation and action history, their dispatch decisions may still be different. Input 2) shows that the execution of the dispatch policy is in a decentralized way, which conditions only on local observations. Last, the dispatch decision of the previous epoch is used as input and GRU cells are adopted as part of the agent network structure, which means dispatch decisions of different epochs within an episode are correlated rather than independent. This design makes QMIX suitable for  solving  multi-step decision making problems, like ours: a sequence of dispatch decisions within an episode should be made for each truck. 

The mixing network $Q_{mix}$ estimates the joint Q-value $Q_{tot}$ as a complex non-linear combination of Q-values for all the $n$ trucks. Also, \textit{HyperNetworks} \cite{DBLP:conf/iclr/HaDL17} is adopted to transform the environment state $s^{k}$ into the parameters of $Q_{mix}$, so the joint Q-value also conditions on the global state information, which has been proved necessary in its original paper \cite{rashid2018qmix}. Based on $Q_{tot}$, the loss function for QMIX can be defined as Equation (5), where $y_{tot}^{k}$ (Equation (6)) is the target joint q-value, which is obtained from the target networks. 
\begin{equation}
Loss(\boldsymbol{\theta}) = \sum_{i=1}^{batchsize}\left[(y_{tot}^{k}-Q_{tot}(\boldsymbol{\tau}^{k},\mathbf{u}^{k}, s^{k}; \boldsymbol{\theta}))_{i}^{2}\right]
\end{equation}

\begin{equation}
y_{tot}^{k} = r^{k}+max_{\mathbf{u}}Q_{tot}(\boldsymbol{\tau}^{k+1},\mathbf{u}, s^{k+1}; \boldsymbol{\theta}^{-})
\end{equation}
After the centralized training, the agent network $Q_{single}$ can be used to give dispatch decisions for each truck based on their individual observation, action history, and agent id in a decentralized way.
\begin{figure}[!htbp]
	\centering
	\includegraphics[width=2.8in,height=1.4in]{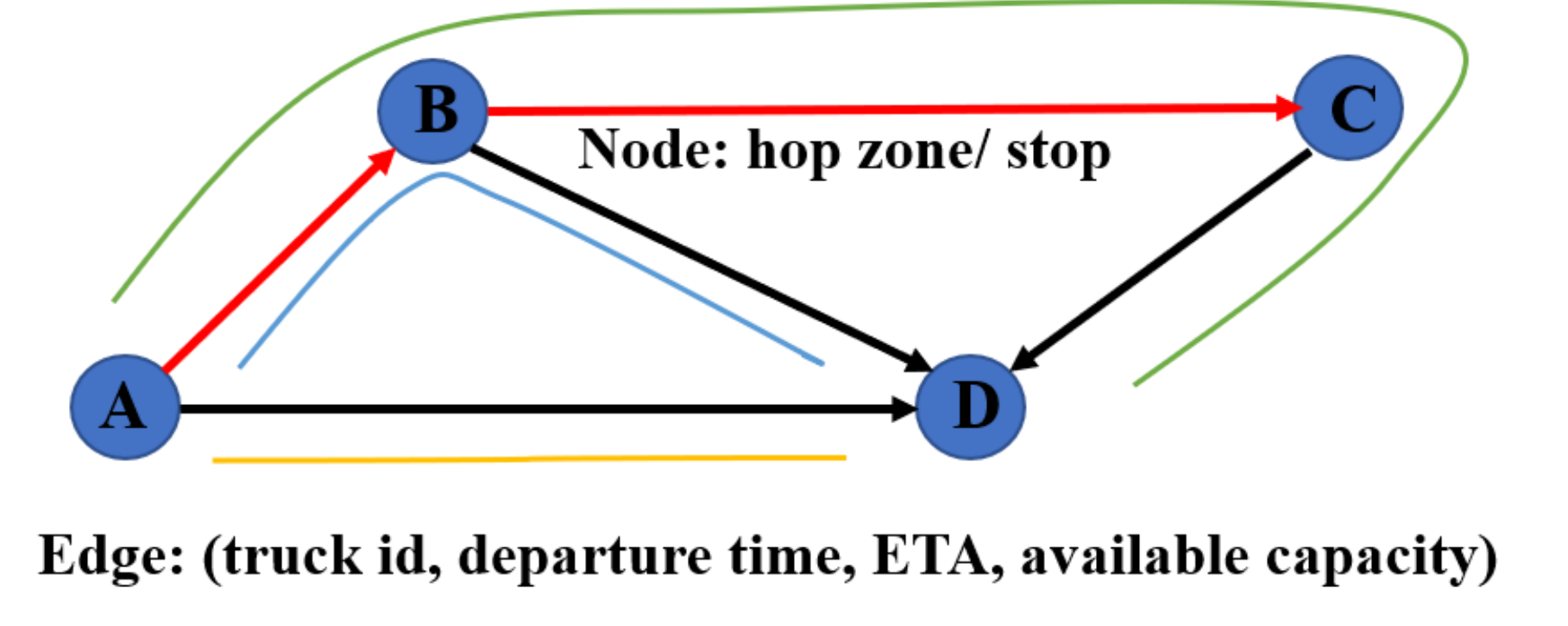}
	\caption{Graph Representation of the Dispatch Decisions}
	\label{3}
\end{figure}

\subsection{Matching Algorithm}\label{subsec:match}

The matching algorithm, shown as \textbf{Algorithm 2}, is executed to match the delivery requests with the dispatch decisions. Considering the number of the packages to deliver is huge, it's nearly impossible to find the optimal matching results for all the packages within a reasonable time. In this case, \textbf{Algorithm 2} adopts a greedy approach to ensure the optimality for every single package (assuming no other packages) and takes the least increase in the total driving time of the fleet as the optimization objective. It mainly includes two parts:

First, in Line 3-14, a weighted directed graph, like Figure \ref{3}, is created based on the dispatch decisions. Each node in the graph represents for a stop that has occurred in the dispatch decisions. While each edge means there is a truck that has been dispatched between the two stops. Also, the id, departure time, and available capacity of the truck and ETA between the two stops are saved with the edge for later use. 

Second, based on the graph, we can adopt DFS (depth-first search) to acquire all the available paths from the source to the destination for each delivery request (Line 16). An available path contains a list of edges and their truck id may be different, which means the package may be delivered by more than one truck (multi-transfer).  In this case, there are some conditions for a path to be available: 1) The package's arrival time at any node in the path should be earlier than the departure time from that node of the truck it transfers to; 2) The package's size should be smaller than that truck's capacity. 

Further, among all the available paths, the one that causes the least increase in the total driving time of the fleet will be chosen (Line 18). For example, there are three available paths (shown in different colors) from $A$ to $D$ in Figure \ref{3}: $A \to D$, $A \to B \to D$, and $A \to B \to C \to D$ and the edges $A \to B$ and $B \to C$ have been matched with other requests. Then, the time cost of the third path should only be the ETA from $C$ to $D$, since it will share the trucks denoted by the red edges with other requests (freight ride-sharing). Similarly, the time cost of the first two paths are respectively the ETA from $A$ to $D$ and ETA from $B$ to $D$. Then, in order to minimize total driving time of the fleet, the path with the least time cost should be chosen. Note that if there are more than one path with the least time cost, the shortest path will be chosen. For example, if the time cost of the three paths in Figure \ref{3} are the same, then the shortest one $A \to D$ will be chosen.

\begin{algorithm}[t]
	\caption{Matching Algorithm}\label{alg:Algo2}
	{
			\begin{algorithmic}[1]
				\State \textbf{Given}: initial capacity of the trucks, dispatch decisions viz. $next\_stop$ for each truck at each valid epoch, request list viz. $(source, destination, size)_{1:N_{r}}$ 
				\State Turn the dispatch decisions into a weighted directed graph as follows (Line 3-14):
				\State Create as many nodes as stops in the dispatch decisions, using the stop id as the node id 
				\State Set the $current\_time$ for each truck as 0
				\For {every truck}
				\For{every valid epoch}
				\State Create a directed edge from $current\_stop$ node 
				\Statex $\qquad\quad$to  $next\_stop$ node
				\State Set the departure time from $current\_stop$ as the 
				\Statex $\qquad\quad$truck's $current\_time$
				\State Acquire the ETA from $current\_stop$ node to 
				\Statex $\qquad\quad$ $next\_stop$ node through Google Map API
				\State Record the id, departure time, ETA, and initial 
				\Statex $\qquad\quad$capacity of the truck as the edge's information
				\State Update the truck's $current\_stop$ as $next\_stop$
				\State Update the truck's $current\_time$ as ETA + 
				\Statex $\qquad\quad$ $current\_time$
				\EndFor
				\EndFor
				\For {every delivery request}
				\State Find all the available paths from $source$ to 
				\Statex $\quad  $ $destination$ using DFS (depth-first search)
				\If{success}
				\State Choose the path that causes the least increase in 
				\Statex $\qquad\quad$ the total driving time of the fleet
				\State Subtract the package size from the available ca-  
				\Statex $\qquad\quad$pacity of the edges it passes through
				\EndIf
				\EndFor
	\end{algorithmic}}
\end{algorithm}

%% file: milp.tex
\section{Integration with MILP}\label{sec:milp}

We further propose a hybrid approach that harnesses DeepFreight and MILP to ensure successful delivery of all the packages. Specifically, experiments show that when most of the requests have been completed and only a small number of packages remain to be optimized, DeepFreight may have unstable training dynamics resulting in undelivered packages (as evidenced in Figure \ref{9:b}). To this end, we leverage MILP to find the exact routing decisions for the small portion of requests that are not efficiently handled by DeepFreight. This leads to an integration of DeepFreight and MILP, denoted as DeepFreight+MILP, shown as Figure \ref{0}. The MILP's formulation and the workflow of DeepFreight+MILP are presented in this section.

\subsection{MILP Formulation}

\noindent\textbf{Parameters:}
\begin{itemize}
	\item $i,j,l\in L=\{0,1,\dots,m-1\}$: index and set of locations;
	\item $k \in N=\{0,1,\dots,n-1\}$: index and set of trucks;
	\item $D$: depot for each truck $k$, a dictionary, $D(k) \in L$;
	\item $t_{i,j} \in \mathbb{N}$: ETA from location $i$ to location $j$, in seconds;
	\item $d_{i,j} \in \mathbb{N}$: delivery demand from location $i$ to location $j$;
	\item $C \in \mathbb{N}^+$: truck capacity, the same for each truck $k$;
	\item $T_{k} \in \mathbb{N}^+$: the time limit for truck $k$.
\end{itemize}

\noindent\textbf{Decision Variables:}
\begin{itemize}
	\item $x_{i,j,k} \in \{0,1\}$: the variable equals $1$, if truck $k$ travels from location $i$ to location $j$, and $0$ otherwise;
	\item $u_{k} \in \{0,1\}$: the variable equals $1$, if truck $k$ is in use, and $0$ otherwise;
	\item $r_{i,j,k} \in \{0,1\}$: the variable equals $1$, if truck $k$ takes the delivery demand from location $i$ to location $j$, and $0$ otherwise;
	
	(Note that a truck takes the path from $i$ to $j$ doesn't mean it will take the delivery request from $i$ to $j$.)
	\item $s_{i,k} \in \mathbb{N}$: the cumulative stop number when truck $k$ arrives at location $i$;
	\item $v_{i,k} \in \mathbb{N}$: the cumulative volume when truck $k$ departs from location $i$;
\end{itemize}

\noindent\textbf{Optimization Objective: }

The optimal solution should minimize the objective function defined as follows, where $T_{1}$ and $T_{2}$ respectively represent the total time use (in hours) and the total volume of the unserved packages, and $\omega_{1}>0, \omega_{2}>0$ are the weights for each term. Based on the results of parameter-adjustment, $\omega_{1}$ is set as 0.5 and $\omega_{2}$ is set as 0.04.
\begin{equation}
target = \omega_{1}T_{1} + \omega_{2}T_{2}
\end{equation}
\begin{equation}
T_{1} = \sum_{i=0}^{m-1}\sum_{j=0}^{m-1}\left[t_{i,j} \times \left(\sum_{k=0}^{n-1}x_{i,j,k}\right)\right]
\end{equation}
\begin{equation}
T_{2} = \sum_{i=0}^{m-1}\sum_{j=0}^{m-1}\left[d_{i,j} \times \prod_{k=0}^{n-1}\left(1-r_{i,j,k}\right)\right]
\end{equation}

\noindent\textbf{Constraints:}
\begin{itemize}
	\item No self-loop:
	\begin{equation}
	\setlength{\abovedisplayskip}{2pt}
    \setlength{\belowdisplayskip}{2pt}
	\forall i \in L, \forall k \in N, x_{i,i,k}==r_{i,i,k}==0
	\end{equation}
	\item Within the time limit:
	\begin{equation}
	\setlength{\abovedisplayskip}{2pt}
    \setlength{\belowdisplayskip}{2pt}
	\forall k \in N, \sum_{i=0}^{m-1}\sum_{j=0}^{m-1}t_{i,j}x_{i,j,k} \leq u_{k} T_{k}
	\end{equation}
	\item Depart from and return to the same depot:
	\begin{equation}
	\setlength{\abovedisplayskip}{2pt}
    \setlength{\belowdisplayskip}{2pt}
	\forall k \in N, \sum_{i=0}^{m-1}x_{i,D(k),k}==\sum_{i=0}^{m-1}x_{D(k),i,k}==u_{k}
	\end{equation}
	\item Form a tour that includes the depot:
	\begin{equation}
	\setlength{\abovedisplayskip}{2pt}
    \setlength{\belowdisplayskip}{2pt}
	\forall k \in N, \forall i \in L, \sum_{j=0}^{m-1}x_{j,i,k}==\sum_{j=0}^{m-1}x_{i,j,k} \leq u_{k}
	\end{equation}
	\item Eliminate any subtour:
	\begin{equation}
	\forall k \in N, s_{D(k),k}==0
	\end{equation}
	\begin{equation}
	\begin{aligned}
	&\forall k \in N, \forall i \in L \backslash \{D(k)\}, \forall j \in L,\\
	&s_{i,k} \geq 1+s_{j,k}-m(1-x_{j,i,k})
	\end{aligned}
	\end{equation}
	\begin{equation}
	\forall k \in N, \forall i \in L, s_{i,k} \geq 0
	\end{equation}
	\item For each delivery demand, one truck can be assigned at most:
	\begin{equation}
	\setlength{\abovedisplayskip}{2pt}
    \setlength{\belowdisplayskip}{2pt}
	\forall i \in L, \forall j \in L, \sum_{k=0}^{n-1}r_{i,j,k} \leq 1
	\end{equation}
	\item If truck $k$ takes the demand from location $i$ to location $j$, truck $k$'s trajectory(tour) should include $i,j$ and $j$ appears after $i$:
	\begin{equation}
	\begin{aligned}
	&\forall i \in L, \forall k \in N, \forall j \in L \backslash \{D(k)\},\\
	&r_{i,j,k} \leq \left(\sum_{l=0}^{m-1}x_{i,l,k}\right)\left(\sum_{l=0}^{m-1}x_{l,j,k}\right),\\
	&r_{i,j,k}(s_{j,k}-s_{i,k}) \geq 0\\
	\end{aligned}
	\end{equation}
	\begin{equation}
	\forall i \in L, \forall k \in N, r_{i,D(k),k} \leq \left(\sum_{l=0}^{m-1}x_{i,l,k}\right)\left(\sum_{l=0}^{m-1}x_{l,D(k),k}\right)
	\end{equation}
	\item Within the capacity limit:
	\begin{equation}
	\setlength{\abovedisplayskip}{2pt}
    \setlength{\belowdisplayskip}{2pt}
	\begin{aligned}
	&\forall k \in N, \forall i \in L \backslash \{D(k)\},\\
	&v_{i,k} == \sum_{j=0}^{m-1}v_{j,k}x_{j,i,k}+\sum_{l=0}^{m-1}r_{i,l,k}d_{i,l}-\sum_{l=0}^{m-1}r_{l,i,k}d_{l,i}
	\end{aligned}
	\end{equation}
	\begin{equation}
	\setlength{\abovedisplayskip}{2pt}
    \setlength{\belowdisplayskip}{2pt}
	\forall k \in N, v_{D(k),k}==\sum_{i=0}^{m-1}r_{D(k),i,k}d_{D(k),i}
	\end{equation}
	\begin{equation}
	\setlength{\abovedisplayskip}{2pt}
    \setlength{\belowdisplayskip}{2pt}
	\forall i \in L, \forall k \in N, 0 \leq v_{i,k} \leq u_{k}C
	\end{equation}
\end{itemize}

Note that MILP's time complexity increases with the number of packages and trucks and is not scalable. In this case, we efficiently combine it with DeepFreight for a scalable solution. 

\subsection{The Workflow of DeepFreight+MILP} \label{subsec:milp+}
 To exploit the advantages of DeepFreight and MILP, we propose an integration of the two, that is DeepFreight gives out dispatch and assignment decisions for most of the requests, while MILP is adopted to find the exact truck routing for the remaining unmatched requests. Its workflow is described as below:

\begin{enumerate}[leftmargin=*]
	\item Get the initial dispatch decisions and matching results using \textbf{Algorithm 1};
	\item Define a key parameter called $efficiency$, which equals the number of packages delivered by the truck divided by its driving time, and calculate $efficiency$ for each truck;
	\item Eliminate all the dispatch decisions of the trucks whose $efficiency$ is lower than the threshold;
	\item Rematch the package list with the pruned dispatch decisions, and get the \textbf{unmatched package list};
	\item Pick the \textbf{new truck list}: choose two trucks from the initial truck list for each distribution center that has unmatched packages, based on the trucks' $priority$ defined as Equation (23); ($available\_time$ means the truck's available time (in seconds) for dispatch before the time limit and $eta$ means the ETA (in seconds) from the truck to the distribution center, so the truck with a higher $priority$ is preferred.)
	\begin{equation}
	priority=available\_time - 2*eta
	\end{equation}
	\item Adopt the MILP optimizer to get the routing result for the \textbf{new truck list} to serve the \textbf{unmatched package list}.
\end{enumerate}

The threshold of $efficiency$ should be determined by the total number of the packages and the total driving time of the fleet that we'd like to achieve. By eliminating the inefficient dispatch decisions, better utilization of the fleet can be realized. However, note that a higher threshold means a larger \textbf{unmatched package list} and the MILP optimizer may not be able to find the optimal routing results when the package number is too large (e.g. shown as Figure \ref{10:a}), so there should be a tradeoff. After fine-tuning, we set the threshold as 0.028 ($\approx40000/(20*20*3600)$) in our experiment, that is how much the $efficiency$ should be if 20 trucks complete 40000 requests with an average driving time of 20 hours.

%% file: evaluations.tex
\section{Evaluation and Results}\label{sec:eval}

In this section, we introduce the setup of the simulator for evaluation, then the implementation details of DeepFreight, and at last, compare the results among the algorithms: DeepFreight, DeepFreight without Multi-transfer, MILP and DeepFreight+MILP.

\subsection{Simulation Setup}\label{subsec:setup}
\begin{figure}
    \centering
    \includegraphics[width=3.2in,height=2.3in]{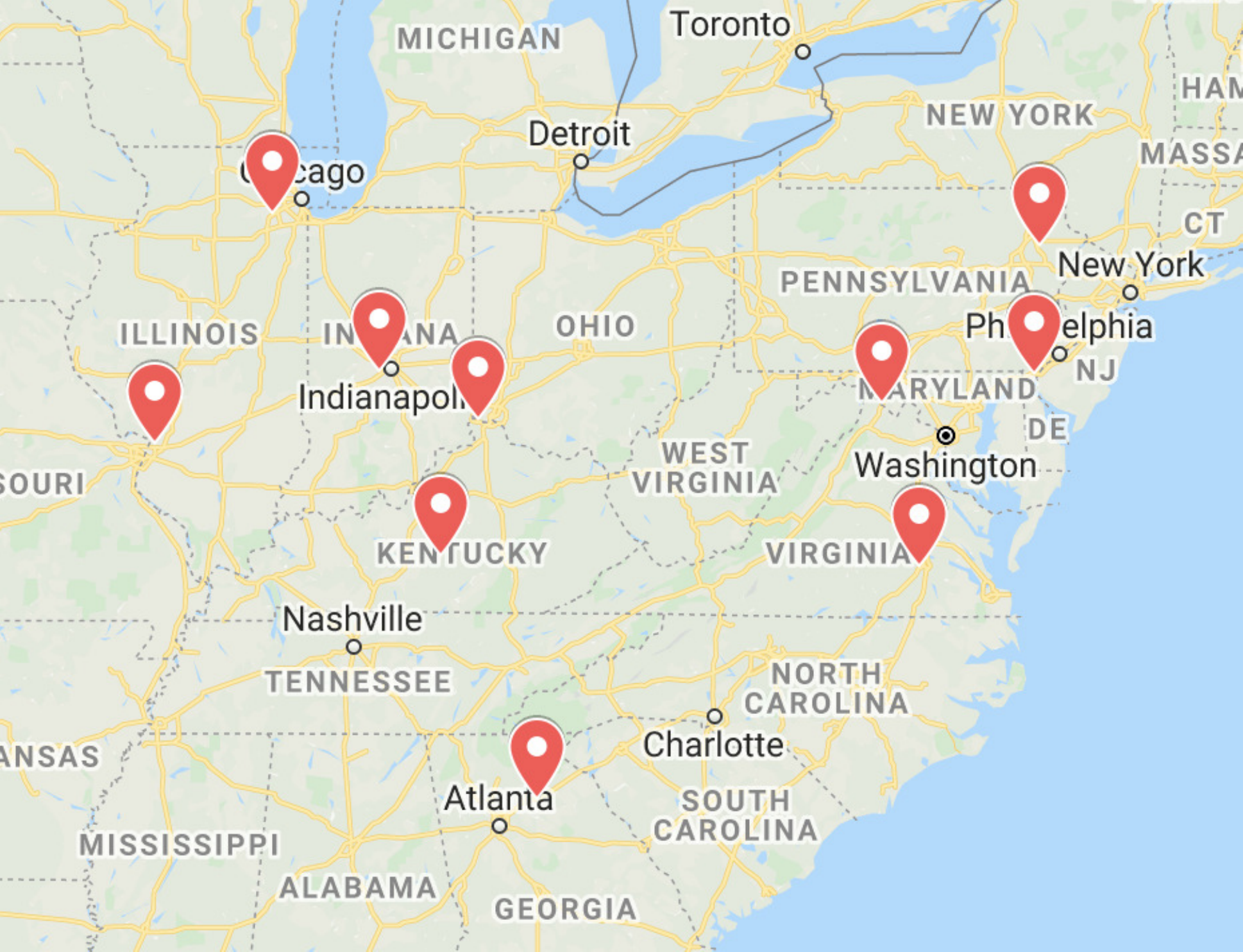}
    \caption{Amazon Distribution Centers on Google Map}
    \label{4}
\end{figure}

As shown in Figure \ref{4}, ten Amazon distribution centers in the eastern U.S. are chosen as the origins and destinations of delivery requests in the simulation. For each episode, $N_t$ trucks should complete $N_r$ randomly generated requests within the time limit $T_{max}$, while trying to minimize the fuel cost. The generation of the request and truck list is model-based, which is introduced as follows:

As mentioned above, each delivery request can be represented as $(source, destination, size)$. Equation (24) shows the distribution of the requests' sources, where $pop_{i}$ represents the population in the vicinity of distribution center $i$ which is acquired through its zip code\footnote{\href{https://www.cdxtech.com/tools/demographicdata/}{https://www.cdxtech.com/tools/demographicdata/}}. After confirming the source of a delivery request, its destination is randomly generated according to the distribution defined as Equation (25), where $norm$ is the normalization coefficient. Further, the size of each package is represented as an integer uniformly distributed between 1 and 30 in our simulation. 

As for the truck list, we need to specify its maximum capacity and its location at the beginning of each operation cycle. In our simulation, an operation cycle includes 7 episodes and the time limit $T_{max}$ for each episode is 2 days. Within an operation cycle, the trucks' locations at the beginning of a new episode are the same as those at the end of the previous episode, while the distribution of the trucks' locations at the first episode is the same as that of the packages' sources (Equation (24)), which is convenient for the fleet to pick up the packages. 

\begin{equation}
    P(source=i)=\frac{pop_{i}}{\sum_{k=1}^{10}pop_{k}}
\end{equation}
\begin{equation}
    P(destination=j | i)=norm * \frac{pop_{j}}{\sqrt{eta_{i,j}}}
\end{equation}

Moreover, we set the truck number $N_t$ as 20, and the maximum capacity for each truck as 30000. Also, with the consideration that trucks should be efficiently used, the total volume of the the delivery requests should match the load capacity of the fleet, so the request number $N_r$ is set as 40000 in our simulation.

\subsection{Implementation Details of DeepFreight}
In this section, we introduce the key parameters about the network structure and training process of DeepFreight.

As previously introduced, the environment state defined in this task includes four items: unfinished delivery requests, distribution/available capacity of the trucks and the current epoch number. The unfinished requests are represented as a matrix, and the element in the $i$-th row and $j$-th column represents the total volume of the packages that need to be delivered from Location $i$ to Location $j$, so the size of this term is $10\times10$ (location number $\times$ location number). The distribution of the trucks means the number of trucks in each location, so its size is 10 (location number). As for the current epoch number, it will be input as a one-hot vector, so its size equals the number of the epochs within an episode, which is set as 10 in our experiment. Compared with the environment space, the individual observation space is for a single truck, so it replaces the trucks' distribution with its current location, which will also be input as a one-hot vector whose dimension is 10 (location number). Moreover, the dimension of the individual action apace is set as 11, where the first 10 dimensions represent the locations to dispatch and the last dimension is an invalid action which means "doing nothing". By now, as shown in Figure \ref{2}, we have known the dimensions of the input and output layers of $Q_{single}$ and $Q_{mix}$. As for the hidden layers, the unit number for all of them are set as 64, and the activation function between the hidden layers are set as ReLu.

During the training process, some tricks about deep Q-learning are adopted. Firstly, we adopt Replay Buffer and Target Network to improve the training stability. The size of the replay buffer is set as 500, which means data of up to 500 episodes can be stored for training use. Moreover, as mentioned in Equation (6), the target network is adopted to calculate the target joint q-value, and the weights of the target network will be copied from $Q_{single}$ and $Q_{mix}$ periodically at an interval of 50 episodes, which can effectively cut down the instability during training. Secondly, Boltzmann Distribution is adopted to improve the exploration in the learning process. The key parameter for this method is $temperature$, and a higher $temperature$ leads to higher exploration. Like the Simulated Annealing Algorithm, the initial $temperature$ is set as 100 and it will drop 0.1 per episode. After 1000 episodes, the agents will begin to use the Greedy Policy instead.

\subsection{Discussions and Results}

\begin{figure}
    \centering
    \includegraphics[width=3.0in]{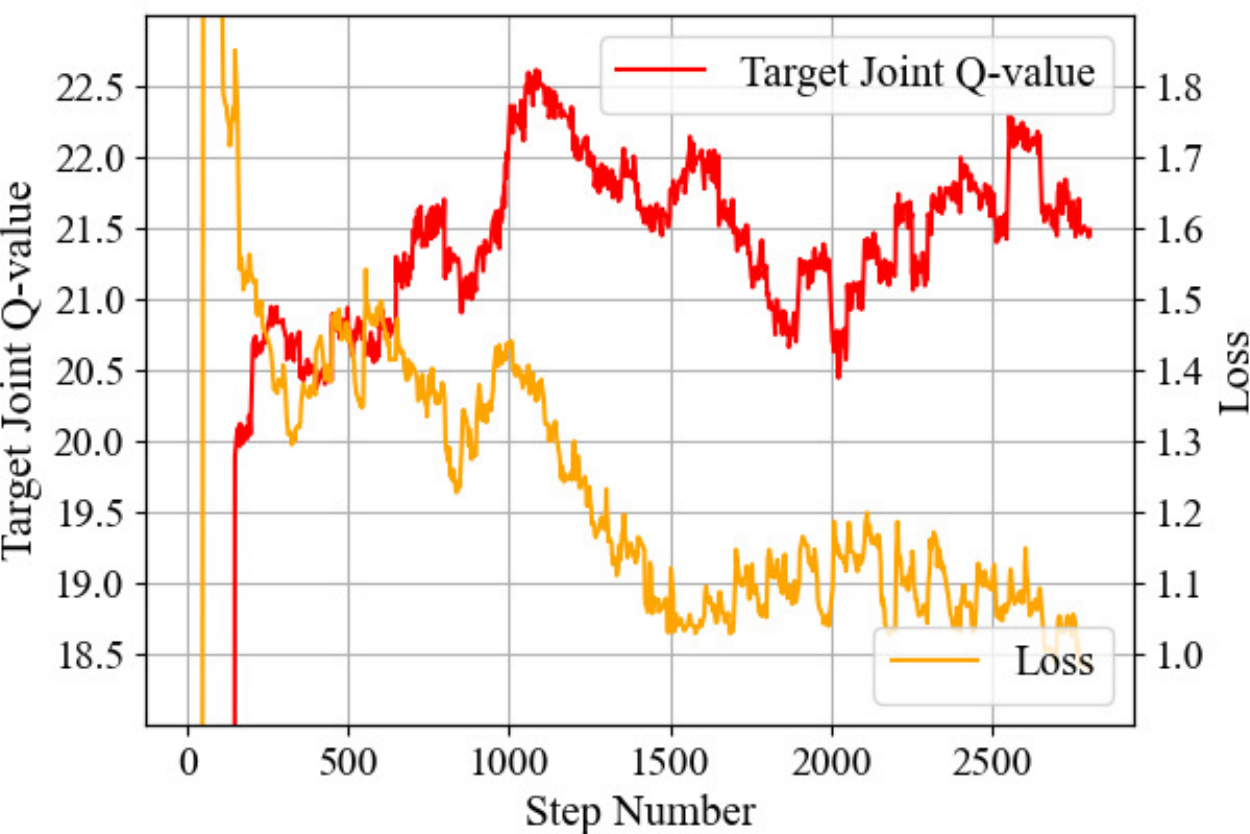}
    \caption{Changing curves of loss function (Equations (5)) and target joint q-value (Equation (6)) for DeepFreight, which show its learning process. We see convergence at approximately 1500 steps and at each step the network is trained for 100 iterations.}
    \label{6}

\end{figure}

\begin{figure*}[t]
\centering
\subfigure[Reward function]{
\label{7:a} 
\includegraphics[width=2.27in]{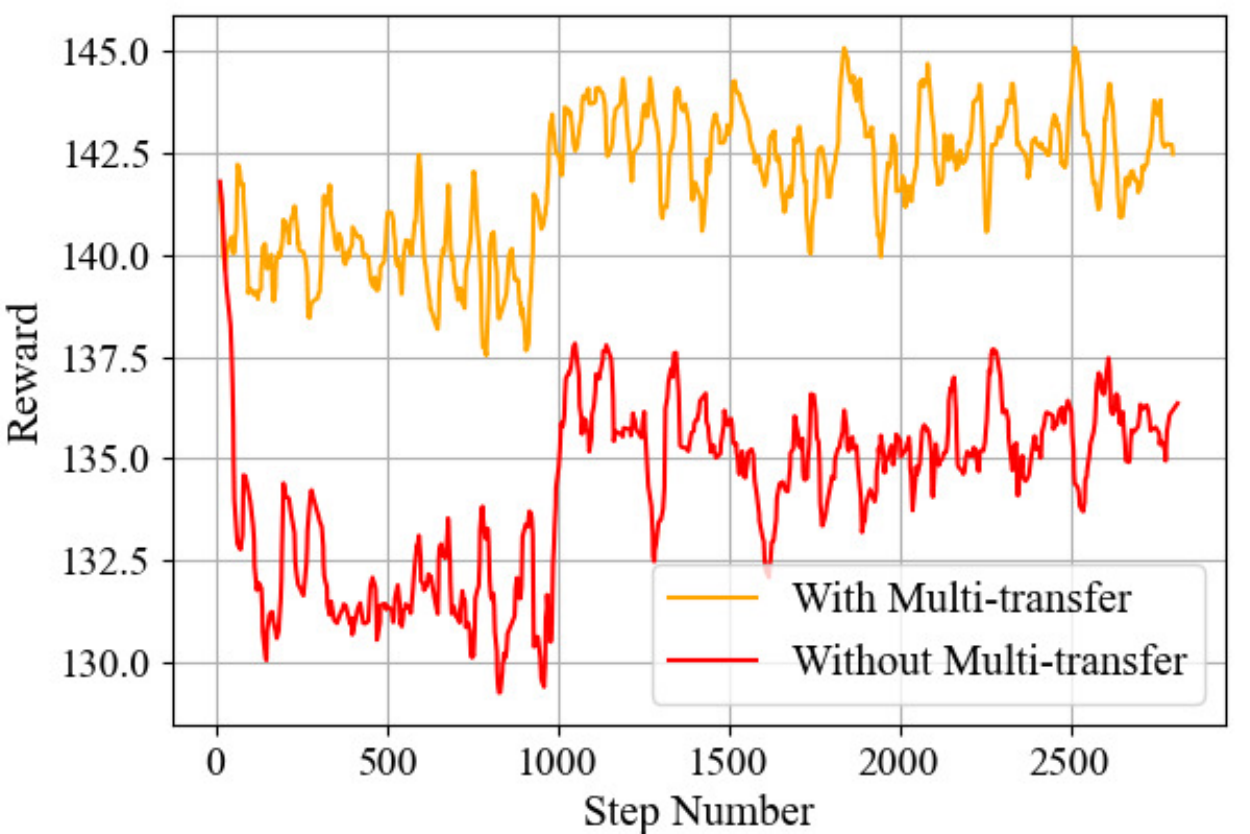}
}
\subfigure[Number of unfinished packages]{
\label{7:b} 
\includegraphics[width=2.27in]{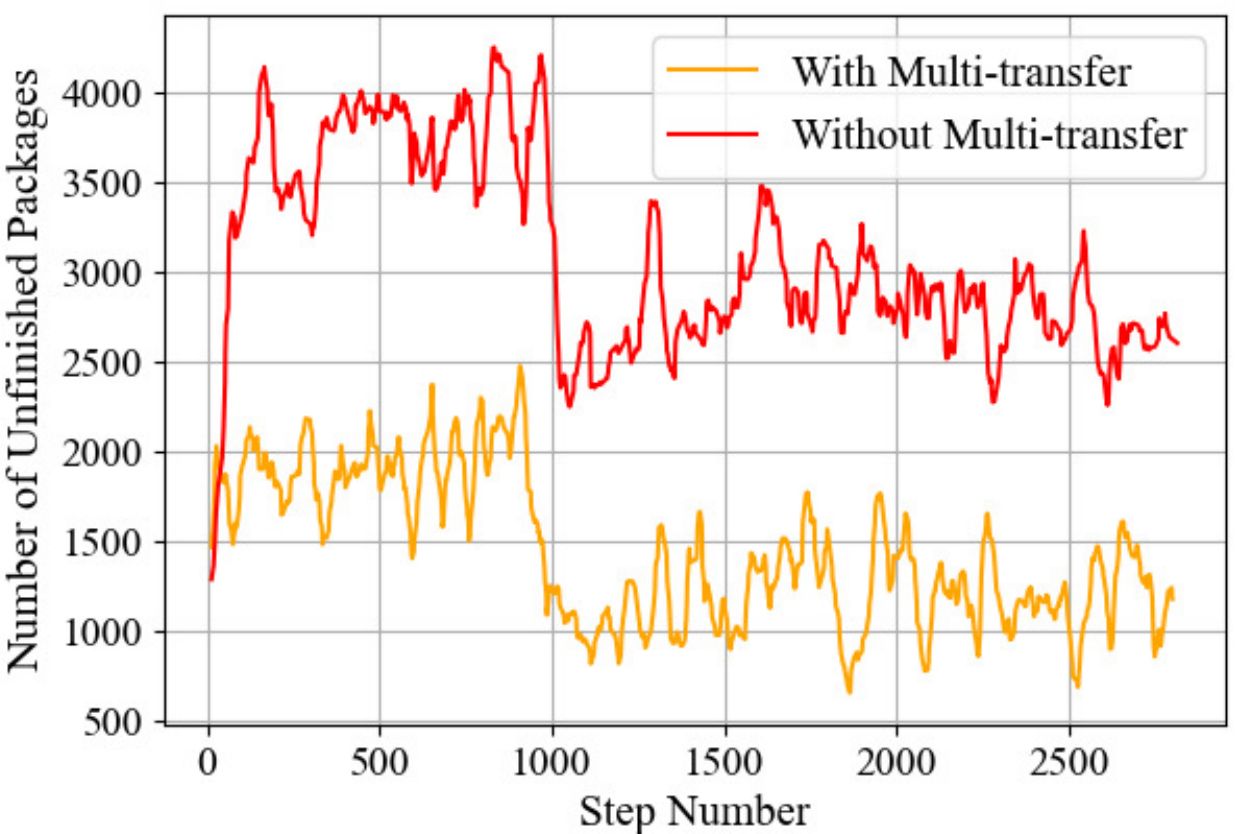}}
\subfigure[Average driving time of the fleet]{
\label{7:c} 
\includegraphics[width=2.27in]{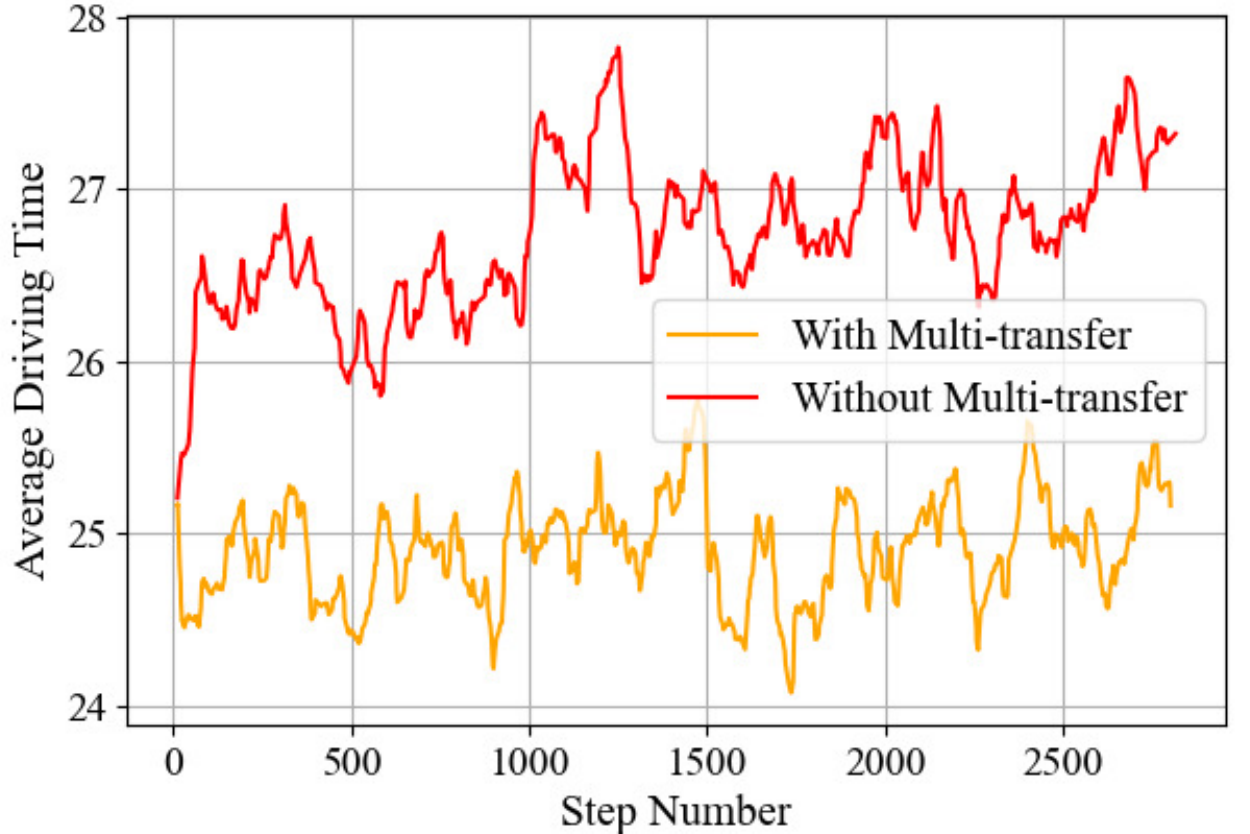}}

\caption{(a)-(c): Training curves of DeepFreight with/without Multi-transfer, in terms of reward function, number of unfinished packages and the fleet's average driving time. DeepFreight with Multi-transfer can complete more package requests with a smaller driving time, and thus has a higher reward, which shows the improvement brought by the flexibility of multiple transfers.}
\label{7} 

\end{figure*}
\begin{figure*}[t]
\centering

\subfigure[Average driving time of the fleet]{
\label{8:a} 
\includegraphics[width=2.27in]{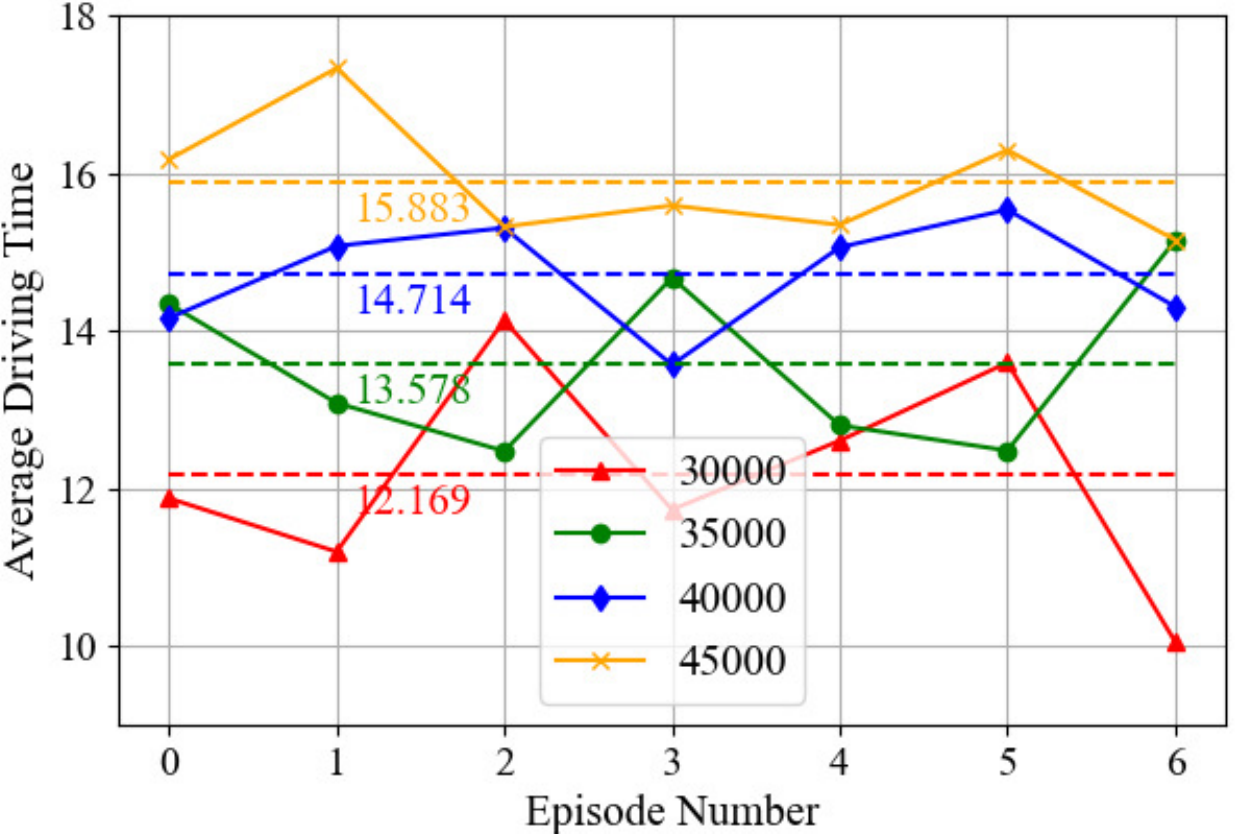}}
\subfigure[Number of unfinished packages]{
\label{8:b} 
\includegraphics[width=2.27in]{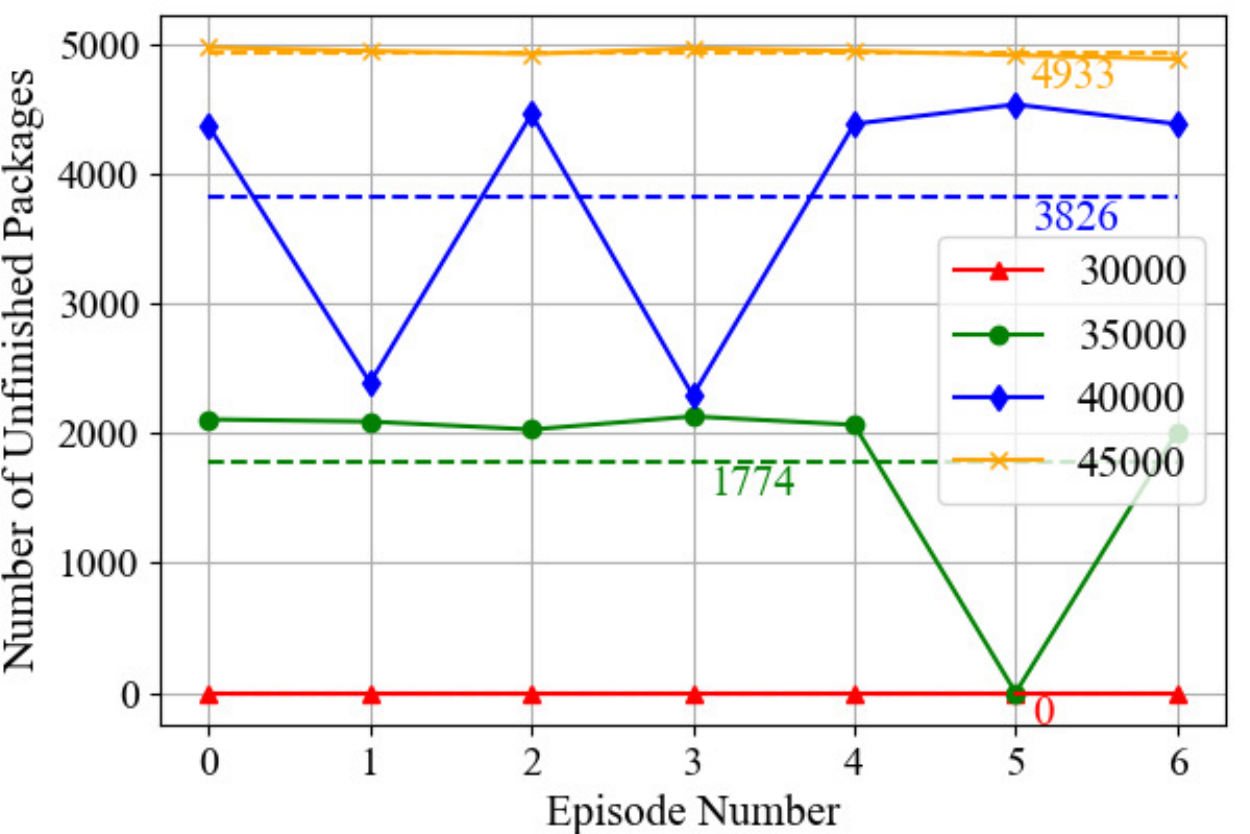}}
\subfigure[Ratio of the unfinished packages]{
\label{8:c} 
\includegraphics[width=2.27in]{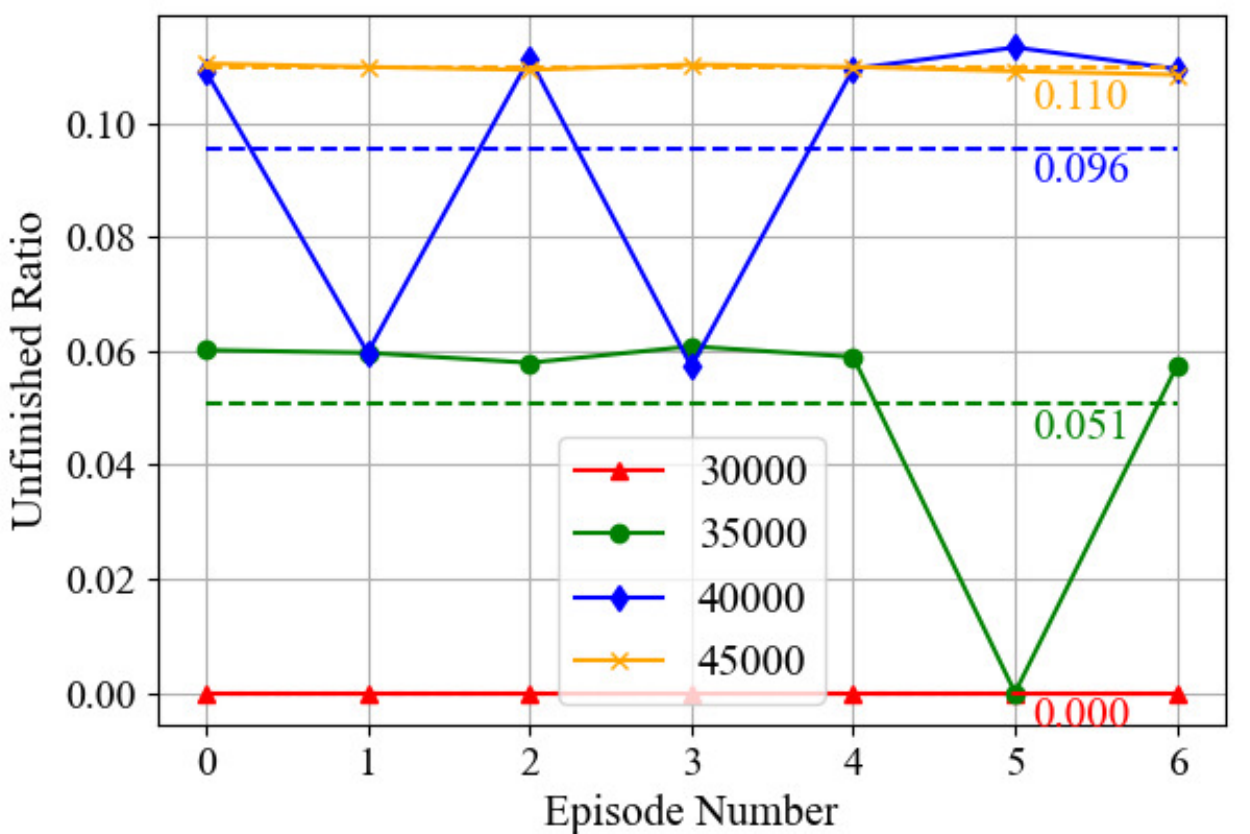}}

\caption{Plotting MILP's performance for varying number of requests (the dotted line: mean value). It can be observed that not only the driving time and the number of unfinished packages increase (shown as (a) (b)), the ratio of unfinished packages (normalized by the total package number) also grows (shown as (c)), demonstrating the poor scalability of MILP.}

\label{8} 
\end{figure*}

\noindent\textbf{Training Process of DeepFreight:}

The training process includes 2800 episodes, that is 400 operation cycles, and at each episode, the networks will be trained for 100 iterations. From the loss curve in Figure \ref{6}, it can be observed that the training process starts to converge at the 1500th episode. After that, the maximum joint Q-value keeps fluctuating around the value 21.5. 

Further, the training effect can be seen from the increase of the reward shown in Figure \ref{7:a}. The reward function for evaluation is the same as that for training (Equation (3)), which is related to the average driving time of the fleet (0-48 hours) and the number of packages served by them (0-40000). $\beta_{1}$ and $\beta_{2}$ are key parameters for defining the reward function. Based on the parameter-adjustment, we set them as: $\beta_1=0.004, \beta_2=0.5$. The increase of the reward is mainly due to the reduction of the unfinished number of packages, which decreases from $\sim$2000 to $\sim$1200 and the lowest number shown in Figure \ref{7:b} is about 700. Also, it's worth noting that the minimum of the unfinished number of packages and average driving time both occur between the 1500th and 2000th episode, so we pick the optimal checkpoint from this interval for the further experiments, as a comparison with other approaches. The comparison is based on their performance within an operation cycle, which equals two weeks as defined in Section \ref{subsec:setup}.

\noindent\textbf{DeepFreight vs. DeepFreight without Multi-transfer:}

In order to show the improvement brought by the flexibility of multiple transfers, we do some comparison between our approach and a freight delivery system without multi-transfer, of which the results are shown in Figure \ref{7:a}-\ref{7:c}. For the system without multi-transfer, each package will be delivered to its destination by a fixed truck, so there is an extra restriction on the truck used when executing the matching policy (\textbf{Algorithm 2}). It can be observed that after convergence, with the flexibility of multi-transfer, the average driving time is shorter ($\sim$2 hours (7.5\%) reduction) and the number of unfinished packages is lower ($\sim$1500 packages (60\%) reduction), which means better utilization of the fleet is realized. For 40000 packages, 53.3\% are delivered with no transfer in the middle, 41.2\%: one hop, 5.5\%: two or more hops, so the extra load/unload effort due to multi-transfer can be overlooked.

\begin{figure*}[t]
\centering
\subfigure[Reward function]{
\label{9:a} 
\includegraphics[width=2.27in]{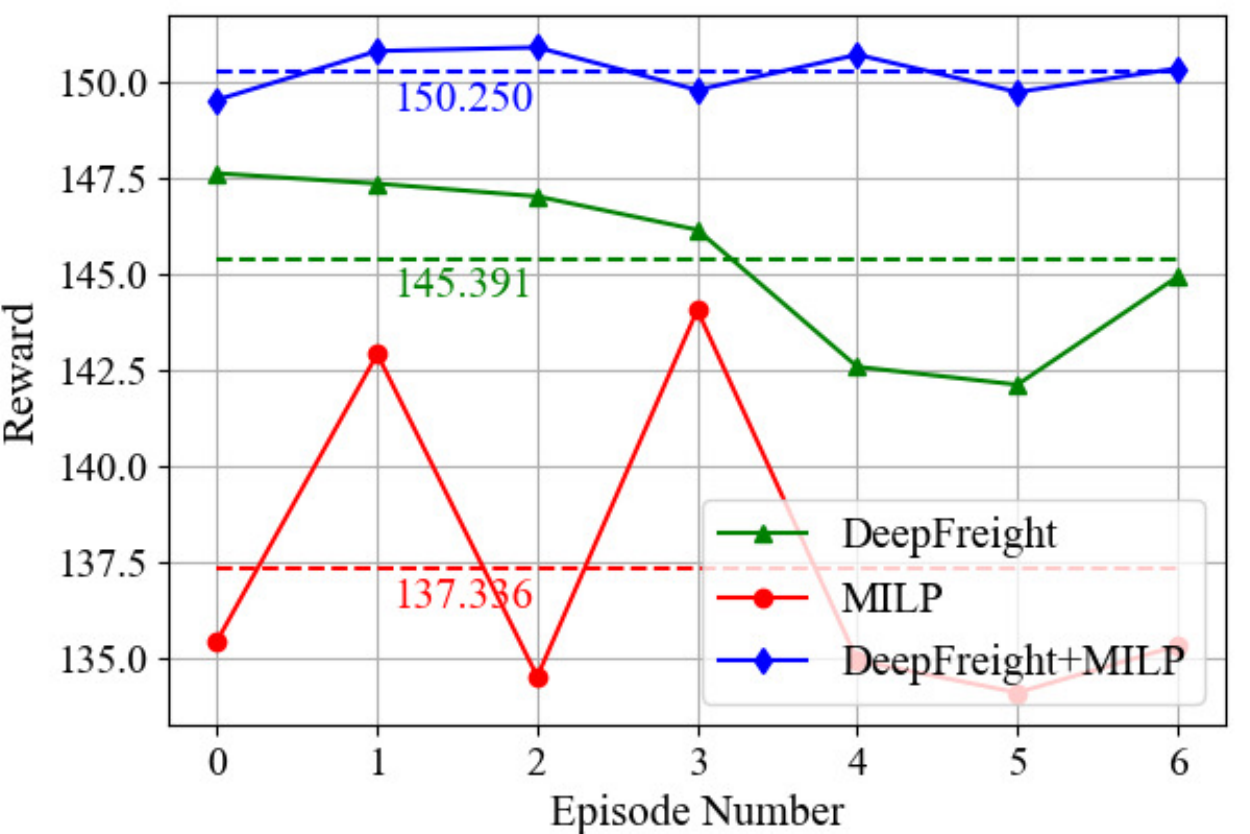}}
\subfigure[Number of unfinished packages]{
\label{9:b} 
\includegraphics[width=2.27in]{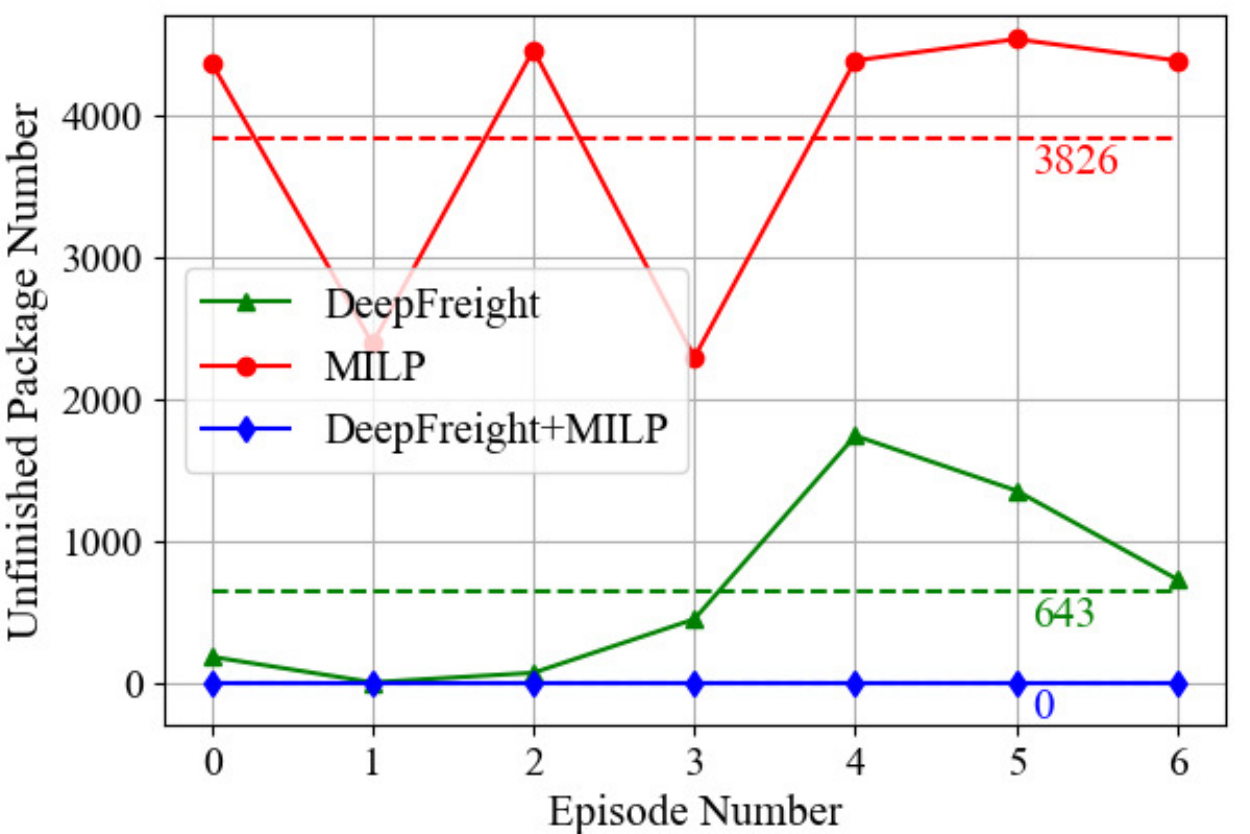}}
\subfigure[Average driving time of the fleet]{
\label{9:c} 
\includegraphics[width=2.27in]{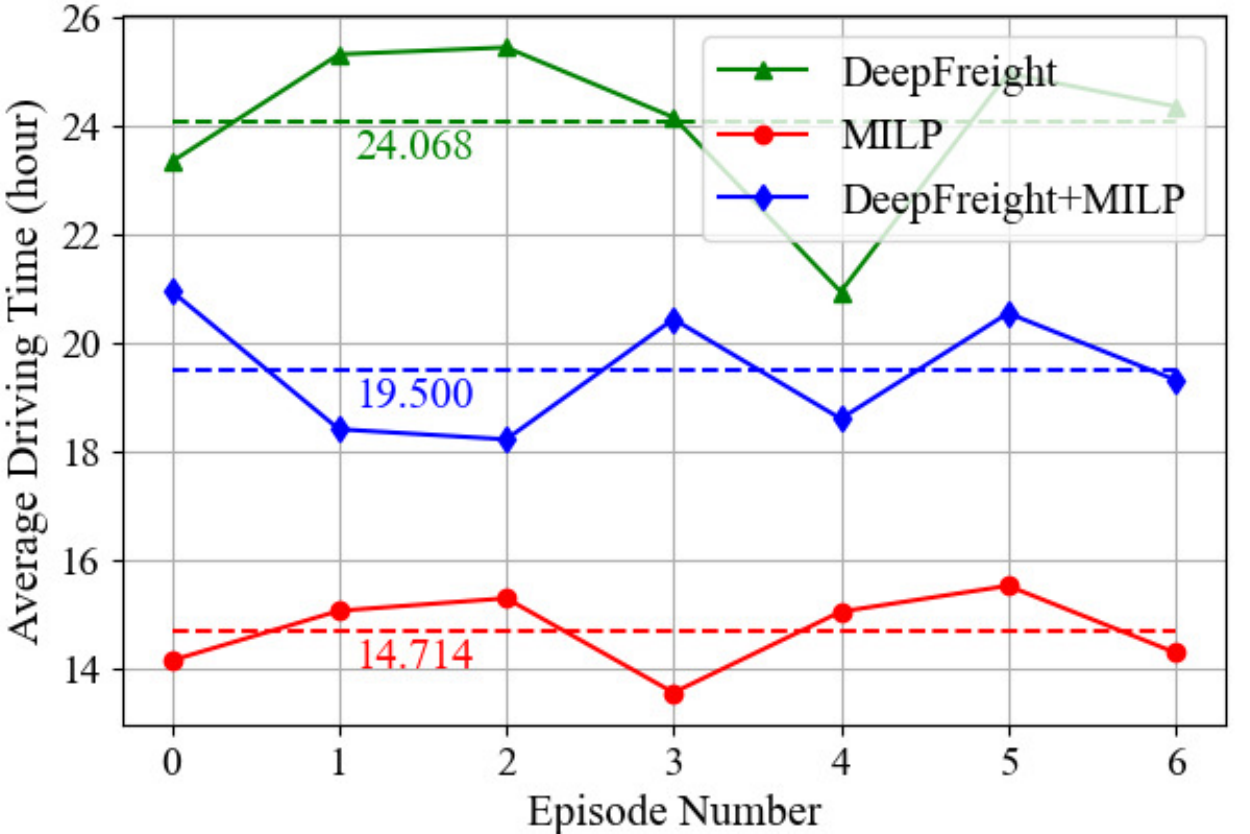}}

\caption{(a)-(c): Comparisons of different algorithms, in terms of the reward function, the number of unfinished packages and the fleet's average driving time (the dotted line: mean value). As a comprehensive index, the reward function we use for evaluation is the same as that for training (defined in Equation (3)), which gives priority to the rate of delivery success. DeepFreight+MILP performs best among these algorithms, since it achieves 100\% delivery success with low fuel consumption.}

\label{9} 
\end{figure*}

\begin{figure*}[!t]
\centering
\subfigure[Optimization objective of MILP]{
\label{10:a} 
\includegraphics[width=2.3in]{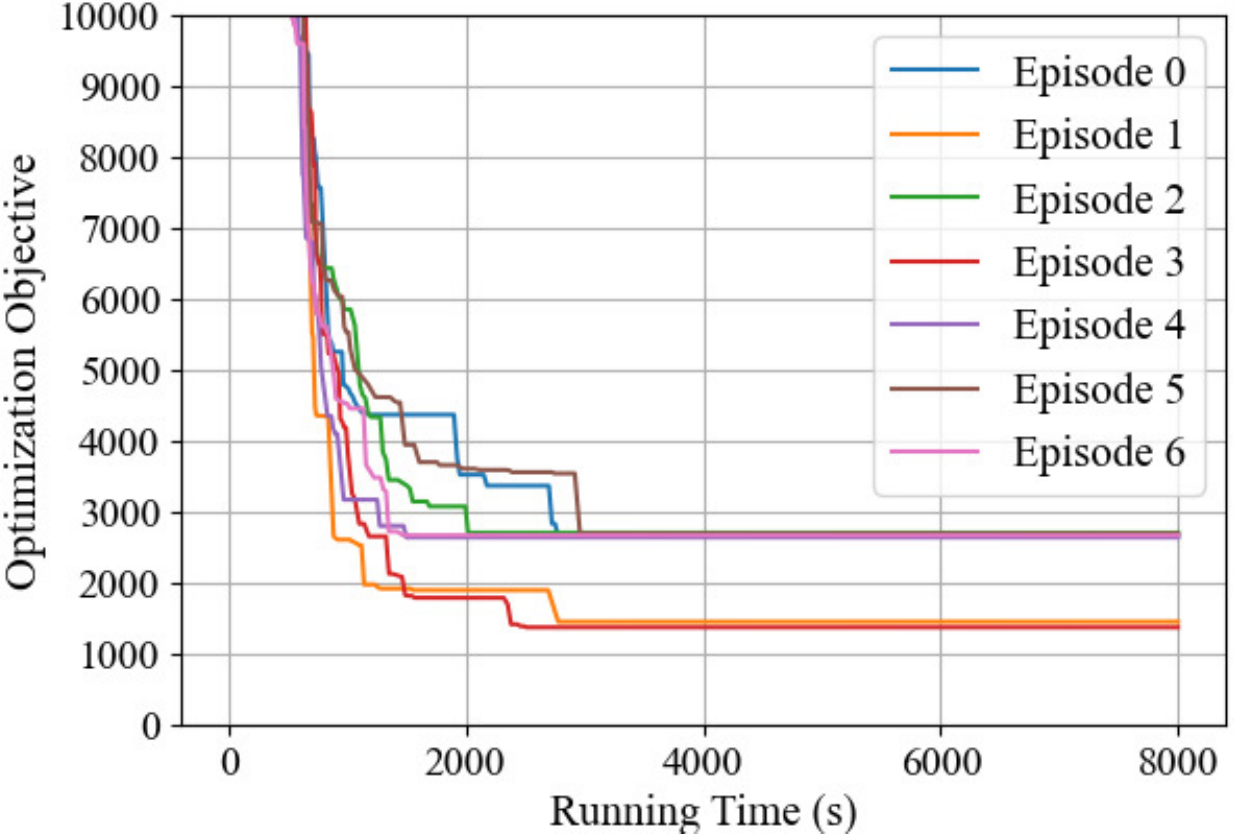}}
\hspace{10mm}
\subfigure[Optimization objective of DeepFreight+MILP]{
\label{10:b} 
\includegraphics[width=2.3in]{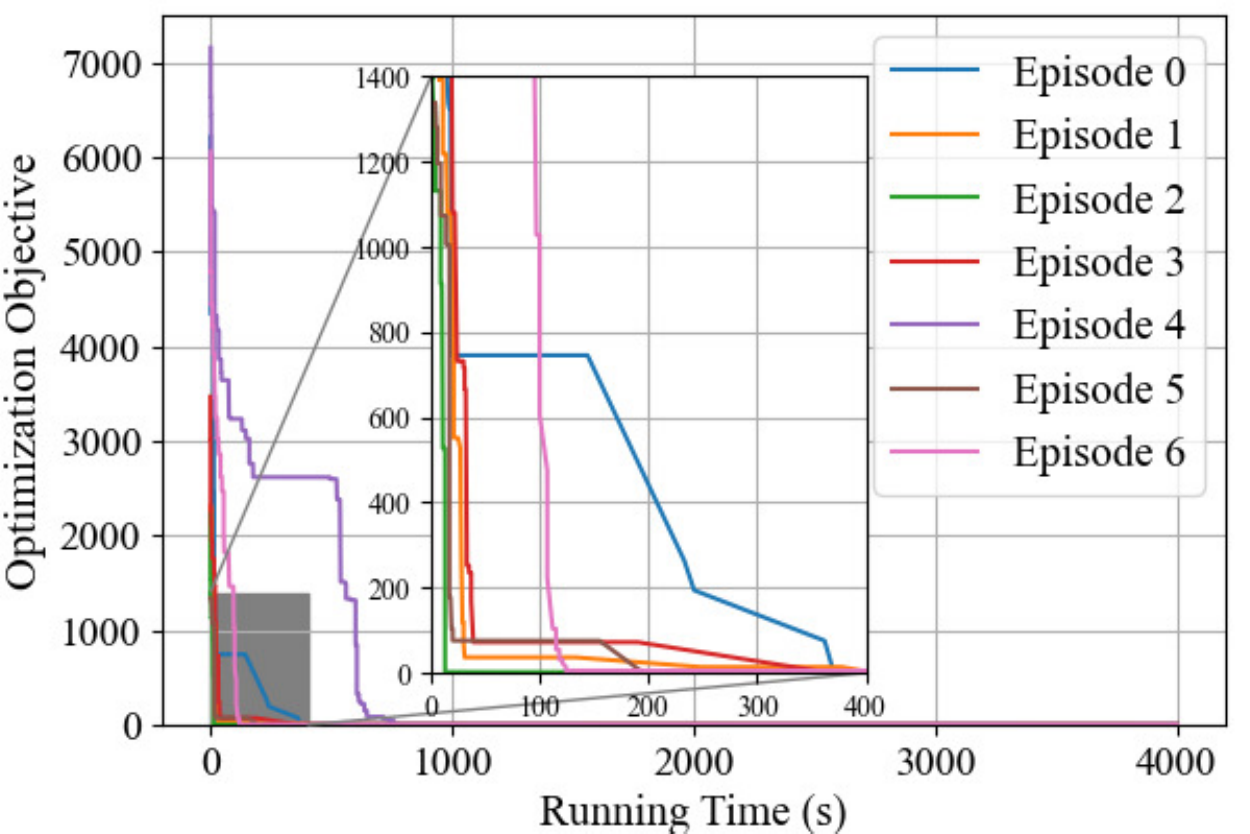}}

\caption{(a)-(b): Convergence curves of the optimization objective of MILP and DeepFreight+MILP over time, when adopting the MILP solver. The  optimization objective is defined in Equation (7)-(9). It can be observed that MILP falls into a local optimum with poor performance, while DeepFreight+MILP can obtain a significantly improved solution within 10 minutes. It demonstrates that DeepFreight+MILP has a better scalability and much lower time complexity.}
\label{10} 
\end{figure*}

\noindent\textbf{DeepFreight vs. MILP:}

MILP is used as the baseline algorithm, which is defined in Section \ref{sec:milp} and solved through a state-of-the-art optimization solver: \textbf{Gurobi} \cite{arulkumaran2017brief}. The evaluation of MILP shows that it has poor scalability. The fleet can carry up to 40000 packages at the same time. Figure \ref{8:a}-\ref{8:c} show that when the number of packages is 30000, the MILP solver can complete all the requests at a fairly low fuel cost (12.169 hours). However, its performance drops dramatically with growing number of packages: both the number and the ratio of the unfinished packages increase. Further, Figure \ref{10:a} shows the convergence curves of the MILP solver, when the number of delivery requests is 40000. It can be observed that the optimization objective (defined as Equation (7)-(9)) converges within 3000s and then the performance doesn't increase any more, which means the MILP solver cannot find a solution for this task even if given more time. Note that the MILP solver runs on a device with an Intel i7-10850H processor.

Next, we compare Deepfreight and MILP. First, in terms of time complexity, $Q_{single}$ trained with DeepFreight can be executed in a decentralized manner. When testing, it can give out the multi-step dispatch decisions for each truck in real time. While, when using MILP, we have to run the optimizer for every episode, since the problem scenario has changed. Also, the time spent will increase with the number of the trucks, distribution centers and packages. Second, as for their performance, Figure \ref{9:b} shows that DeepFreight can complete more requests than MILP. However, its performance is unstable, which is one of the characteristics of reinforcement learning. Also, the average driving time taken of DeepFreight is much higher than that of MILP. 

Overall, as compared to MILP, DeepFreight is transferable among different problem scenarios (different distributions of the trucks and delivery requests) and has lower time complexity, but the average driving time and number of unfinished packages  need to be further reduced, which motivates our design of DeepFreight+MILP.

\noindent\textbf{DeepFreight vs. DeepFreight+MILP:}

Among all the evaluation metrics, serving all the packages is given the top priority. As shown in Figure \ref{9:b}, DeepFreight+MILP completes all the delivery requests within an operation cycle, eliminating the uncertainty and instability of DeepFreight, which makes it suitable for the industrial use. Figure \ref{9:c} shows that DeepFreight+MILP has a further reduction in average driving time as compared with DeepFreight and with comparable driving time as MILP, DeepFreight+MILP realizes a 100\% delivery success. Figure \ref{10:b} shows the convergence curves of DeepFreight+MILP when using the optimizer, and it can be observed that we can get the routing result for the rest of the packages within 10 minutes. That is because most of the packages have been served by the dispatch decisions made by DeepFreight and only 4000-5000 packages are left for MILP to serve. Also, not all the trucks are used for dispatch -- as defined in Section \ref{subsec:milp+}, only the trucks that are closer to the packages' origins and have more available driving time for dispatch before the time limit are chosen as the new truck list. In this case, DeepFreight+MILP can still be adopted when the number of packages or trucks is large, which ensures its scalability.

Overall, DeepFreight+MILP performs best among these algorithms, because it not only has better scalability and lower time complexity but also can ensure a 100\% delivery success with fairly low fuel consumption.

%% file: conclusion.tex
\section{Conclusion}\label{sec:conc}
This paper proposes DeepFreight, a novel model-free approach for multi-transfer freight delivery based on deep reinforcement learning. The problem is sub-divided into truck-dispatch and package-matching, where QMIX, a multi-agent reinforcement learning algorithm, is adopted for training the dispatch policy and Depth First Search is used for matching in a greedy fashion. This approach is then integrated with MILP for further optimization. The evaluation results show superior scalability and improved performance of the combined system as compared to the MILP formulation alone and the learning-based solution alone.

Extending the work with different priority packages is an important future direction. In this case, real-time decisions for accepting priority packages within the day could also be investigated.